\documentclass[preprint,12pt]{elsarticle}

\usepackage{amssymb}
\usepackage{amsmath, epsfig,amssymb,dsfont}
\usepackage{subfigure}
\usepackage{amsthm}
\usepackage{booktabs}
\usepackage{multirow}           

\journal{Signal Processing}

\begin{document}

\begin{frontmatter}



\title{Enhancing Action Recognition from Low-Quality Skeleton Data via Part-Level Knowledge Distillation}

 \author[1]{Cuiwei Liu}
 \author[1]{Youzhi Jiang}
 \author[2]{Chong Du}
 \author[1]{Zhaokui Li}
 \affiliation[1]{organization={Shenyang Aerospace University},
             city={Shenyang},
             postcode={110136},
             country={China}}

 \affiliation[2]{organization={Shenyang Aircraft Design and Research Institute},
             city={Shenyang},
             postcode={110034},
             country={China}}

\begin{abstract}

Skeleton-based action recognition is vital for comprehending human-centric videos and has applications in diverse domains. 
One of the challenges of skeleton-based action recognition is dealing with low-quality data, such as skeletons that have missing or inaccurate joints.
This paper addresses the issue of enhancing action recognition using low-quality skeletons through a general knowledge distillation framework. 
The proposed framework employs a teacher-student model setup, where a teacher model trained on high-quality skeletons guides the learning of a student model that handles low-quality skeletons. 
To bridge the gap between heterogeneous high-quality and low-quality skeletons, we present a novel part-based skeleton matching strategy, which exploits shared body parts to facilitate local action pattern learning. 
An action-specific part matrix is developed to emphasize critical parts for different actions, enabling the student model to distill discriminative part-level knowledge.
A novel part-level multi-sample contrastive loss achieves knowledge transfer from multiple high-quality skeletons to low-quality ones, which enables the proposed knowledge distillation framework to include training low-quality skeletons that lack corresponding high-quality matches.
Comprehensive experiments conducted on the NTU-RGB+D, Penn Action, and SYSU 3D HOI datasets demonstrate the effectiveness of the proposed knowledge distillation framework.

\end{abstract}


\end{frontmatter}


\section{Introduction}	
Action recognition is a crucial step toward human-centric video understanding and has widespread applications in video surveillance, human-machine interaction, virtual reality, autonomous driving, and health care.
Action executions can be represented in various modalities, including RGB videos, RGB+Depth videos, optical flows, and skeleton data in the form of 2D/3D coordinates of human joints.
Among them, skeleton data can accurately express the subtle movements of joints and appear robust against complicated background, light changes, and body scale variations~\cite{ren2020survey}.
Besides, skeleton data are significantly more compact than other modalities and thus less time-consuming for action recognition as well as data transmission in practical applications.
Along with the advance of pose estimation algorithms and depth sensors, skeleton data are more easily accessible and skeleton-based action recognition has gained increasing attention in recent years.

\begin{figure}[t]
	\begin{center}
		\vskip -0pt
		\includegraphics[width=0.7\textwidth]{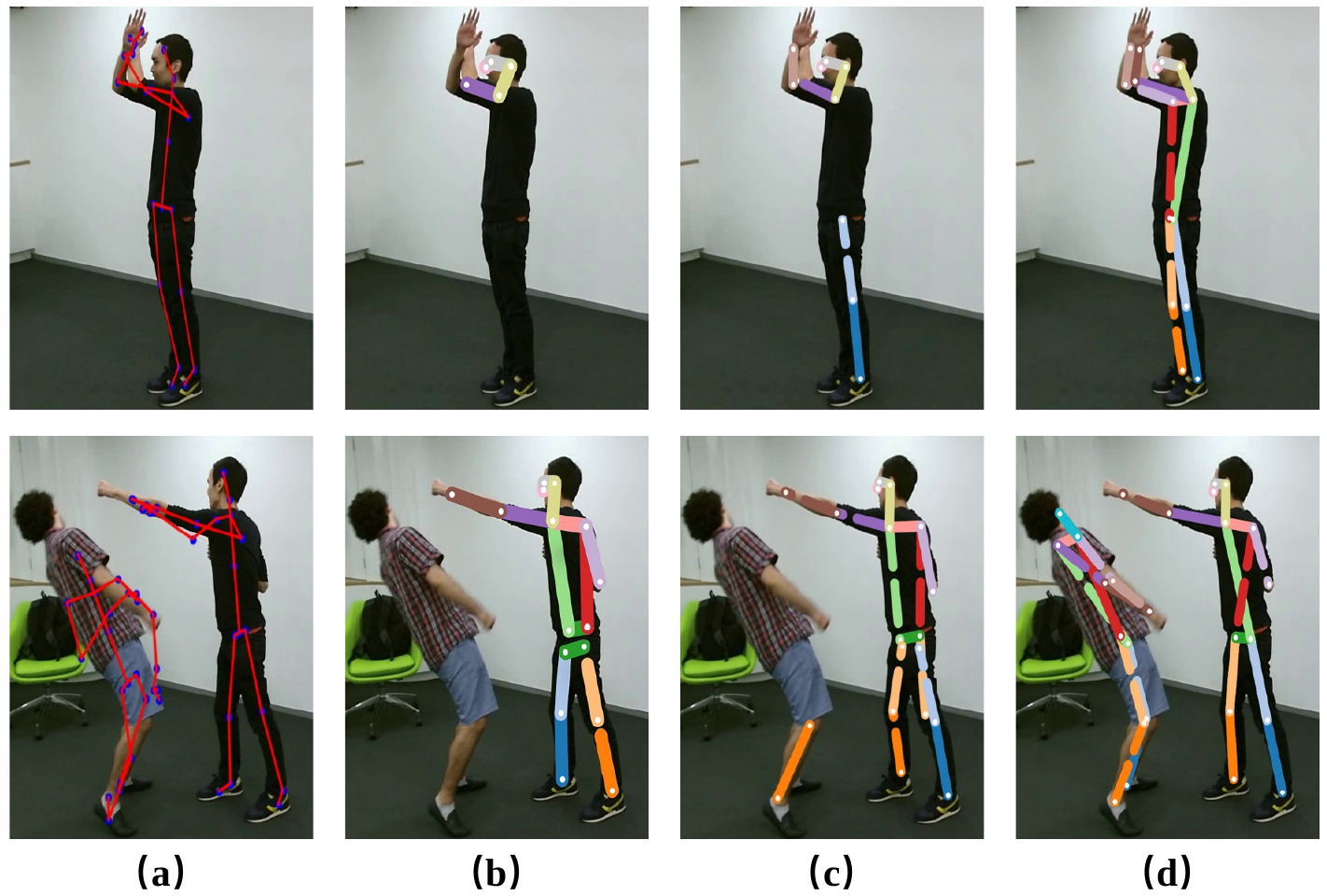}
	\end{center}
	\vskip -15pt
	\caption{Visualization of skeleton data. (a) 3D poses generated by Microsoft Kinect V.2. (b) 2D poses estimated by PifPaf~\cite{kreiss2019pifpaf} with MobileNetv3Small~\cite{howard2019searching} as the backbone. (c) 2D poses estimated by PifPaf with ShufflfleNetv2x1~\cite{ma2018shufflenet} as the backbone. (d) 2D poses estimated by PifPaf with ResNet152~\cite{he2016deep} as the backbone.}
	\vskip -15pt
	\label{fig:data}
\end{figure}

Skeleton extraction is a prerequisite for skeleton-based action recognition and has a significant influence on the final action recognition accuracy.
Depth sensors, such as Microsoft Kinect~\cite{zhang2012microsoft}, can collect 3D coordinates of joints in real-time by using the scene depth information but are easily affected by natural light in outdoor applications.
Human pose estimation algorithms~\cite{cao2017realtime,kreiss2019pifpaf} calculate 2D coordinates of human skeleton joints on top of RGB videos and can be applied to extensive scenarios as most videos are captured by RGB cameras.
As demonstrated in the early study~\cite{2022Revisiting}, modern pose estimators employ strong backbones for feature extraction and can provide 2D poses of better quality than 3D poses extracted from depth sensors at the cost of computational efficiency.
However, computational efficiency is a crucial requirement to achieve fast system reaction in many real-world applications.
In this case, it is desirable to use faster pose estimators that lower the computation consumption but sacrifice accuracy, thereby leading to low-quality skeletons including more inaccurate or missing joints as shown in Fig.~\ref{fig:data}.
Actually, pose estimation noise is also inevitable in high-quality skeletons owing to the partial occlusion of human body and the limited view.
Most of the previous skeleton-based action recognition models~\cite{shahroudy2016ntu,ke2017new,yan2018spatial} are learned and evaluated on high-quality skeletons and assume that mild skeleton noises can be implicitly solved by the networks.
As a result, the action recognition performance drops significantly when applying these models on low-quality skeletons with intensive noises.

\begin{figure}[t]
	\begin{center}
		\vskip -0pt
		\subfigure[]{\includegraphics[width=0.28\textwidth]{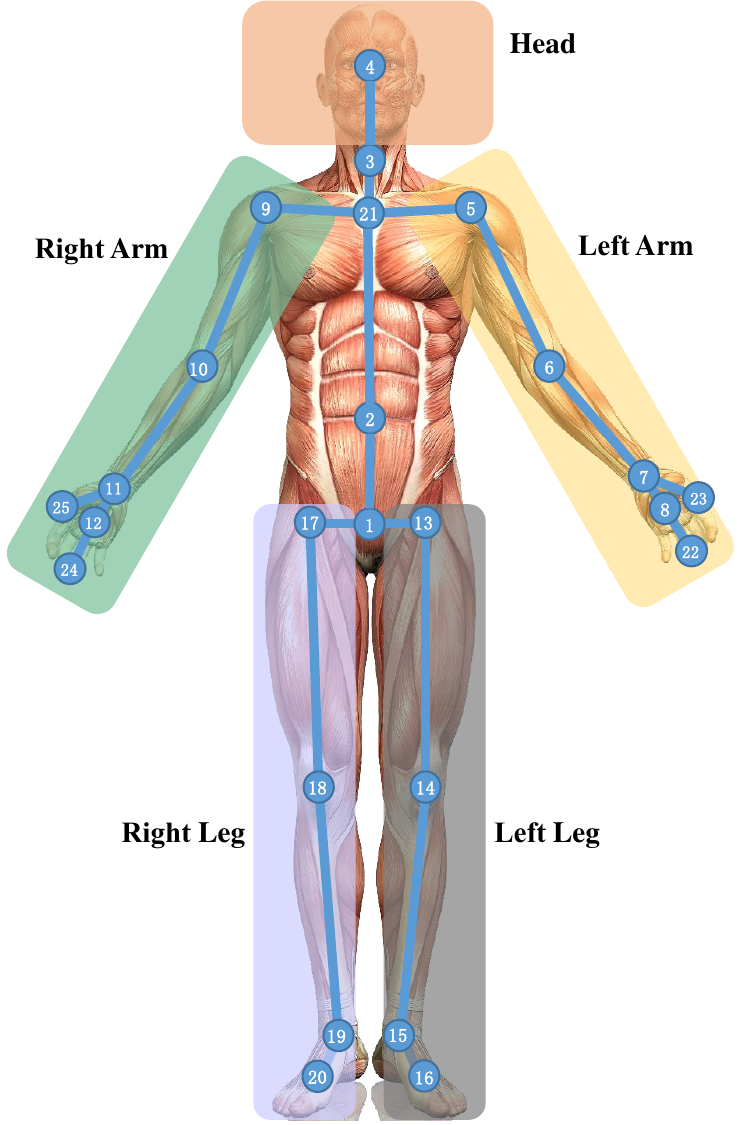}}
		\subfigure[]{\includegraphics[width=0.28\textwidth]{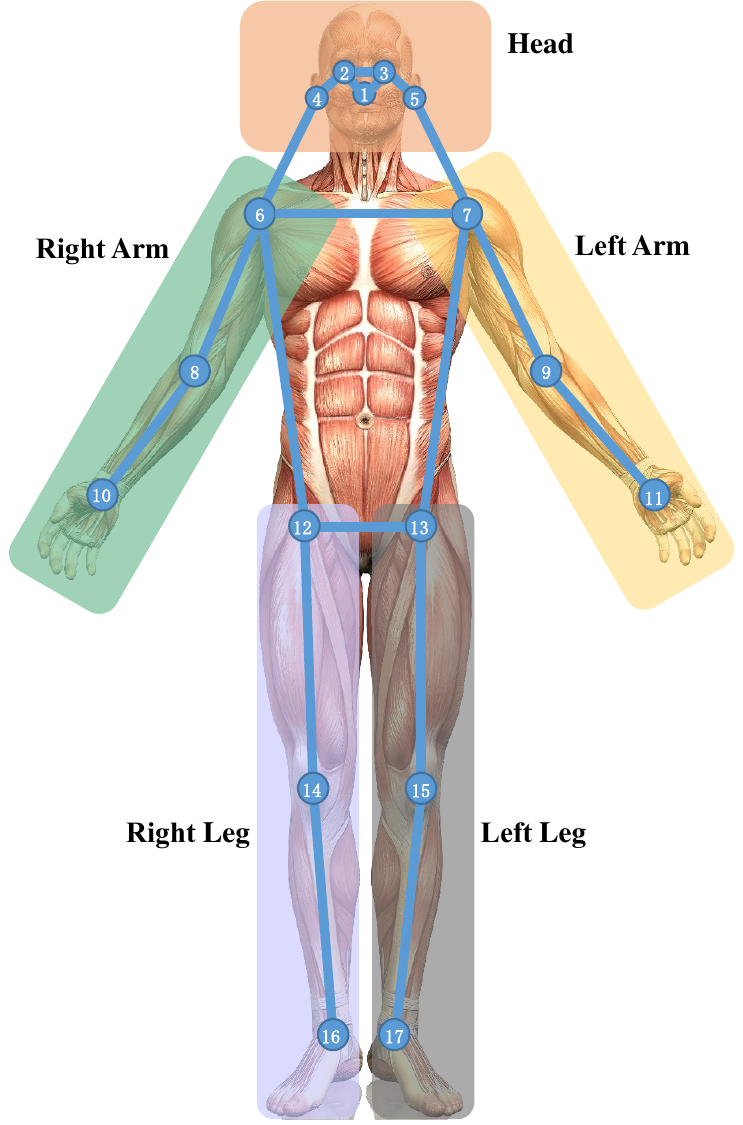}}
	\end{center}
	\vskip -20pt
	\caption{Visualization of two skeleton graphs. (a) The skeleton graph produced by Microsoft Kinect V.2 contains 25 joints. (b) The skeleton graph obtained by the PifPaf pose estimator~\cite{kreiss2019pifpaf} contains 17 joints.}
	\vskip -10pt
	\label{fig:part}
\end{figure}

A few works~\cite{liu2017enhanced,nie2019view,demisse2018pose} take low-quality skeletons into account and explicitly perform skeleton denoising operations as data augmentation.
Although the denoising operations can revise some big errors in joint coordinates through filters or prior of body structure, it is intractable to develop a general skeleton denoising method as the noise patterns are unclear in realistic scenarios.
Instead of explicit skeleton denoising process, some works~\cite{song2020richly,yoon2022predictively,song2022learning,bian2021structural} resort to learning noise-robust representations from raw skeletons.
Towards this goal, ensemble learning is utilized to construct multi-stream models~\cite{song2020richly} at the cost of more parameters and lower model efficiency.
Another strategy~\cite{yoon2022predictively,song2022learning,bian2021structural} is to transfer underlying knowledge from high-quality skeletons to enhance representations of low-quality skeletons.
The studies in~\cite{yoon2022predictively,song2022learning} perform knowledge transfer by using global features of all joints, while~\cite{bian2021structural} aligning the features of low-quality skeletons and their corresponding high-quality ones at joint-level.
However, we argue that there are two issues pertaining to the above methods.
\begin{itemize}
	\item 
	Joint-level knowledge transfer~\cite{bian2021structural} can better deal with skeleton noise compared to global knowledge transfer~\cite{yoon2022predictively, song2022learning}, but cannot be applied to high-quality and low-quality skeletons with heterogeneous pose graphs.
	Poses produced by different pose estimators or depth sensors may differ a lot in joints and links as shown in Fig.~\ref{fig:part}, which makes it hard to perform joint-to-joint feature alignment.
	\item 
	Knowledge transferred from high-quality skeletons is not guaranteed to be discriminative for action recognition, since the intrinsic relationships between joints and actions are neglected.
	Actually, the effects of joints on identifying various actions are rather different. 
	For example, occluded legs are trivial for action ``drinking water" but make it difficult to identify action ``hopping".
\end{itemize}

This paper proposes a general knowledge distillation framework which enhances action recognition based on low-quality skeletons by transferring fine-grained knowledge from heterogeneous high-quality skeletons.
A teacher model is pre-trained on high-quality skeletons to guide the learning of a student model which takes low-quality skeletons as input. 
In order to achieve knowledge transfer between heterogeneous high-quality and low-quality skeletons, we develop a novel part-based skeleton matching strategy. 
While heterogeneous skeleton graphs represent the human body with different joints and links, they often share common body parts, such as the head, arms, and legs (see Fig.~\ref{fig:part}).
Based on these observations, we utilize body parts to capture local patterns of actions and enforce the student model to imitate part representations produced by the teacher model.
As body parts make different contributions to recognizing various actions, we develop an action-specific high-efficiency part matrix to emphasize key parts that are important for each action.
Consequently, the student model can acquire discriminative part-level knowledge that is crucial for accurate action identification.
In real life, low-quality skeletons are commonly inaccurate or even incomplete due to self-occlusion caused by the deformation of human body and external occlusion caused by the human-object interaction.
With the proposed part-based skeleton matching strategy, the student model can obtain fine-grained guidance from high-quality skeletons to compensate the lost local information or rectify the inaccurate joints.
Unlike previous works~\cite{yoon2022predictively,bian2021structural} that require one-to-one match between a low-quality skeleton sequence and a high-quality skeleton sequence, we introduce a part-level multi-sample contrastive loss to distill action knowledge from multiple high-quality skeleton sequences for a low-quality skeleton sequence.
Thus the proposed method also incorporates action instances with only low-quality skeletons into training, distilling reasonable knowledge from high-quality skeletons of other instances.

Our main contributions are summarized as follows:
\begin{itemize}
	\item We propose a general knowledge distillation framework for action recognition based on low-quality skeletons. The proposed framework distills discriminative knowledge from heterogeneous high-quality skeletons to enhance representations of low-quality skeletons with intensive noise.
	\item We develop a novel part-based skeleton matching strategy to accomplish part-level knowledge transfer between heterogeneous skeleton graphs. An action-specific high-efficiency part matrix is explored to adaptively distill discriminative information from high-quality skeletons according to the characteristics of different actions.
	\item We introduce a novel part-level multi-sample contrastive loss to achieve intra-class knowledge transfer, which makes the proposed knowledge distillation method also available to individual training low-quality skeletons with no corresponding high-quality matches.
\end{itemize}

\section{Related Work}
\subsection{Skeleton-based action recognition}

Most of the advanced skeleton-based action recognition methods leverage the powerful learning capabilities of deep neural networks.
Among them, some works~\cite{shahroudy2016ntu,zhang2017view,ng2021multi} adopt Recurrent Neural Networks (RNN) due to their good capability for capturing temporal dynamics of actions.
Another group of works~\cite{2022Revisiting,ke2017new,xia2021laga} transform coordinates of skeleton joints to pseudo-images or skeleton heat maps, which are then fed to Convolutional Neural Networks (CNN) to learn discriminative features.
In general, CNN-based methods are effective on exploiting local relationships among multiple joints while RNN-based methods do well in long-term temporal modeling.
Therefore, some methods~\cite{zhang2019view,li2021memory} integrate CNN and RNN into a unified framework and make the final decision by fusing their predictions.
The recent development of Graph Convolutional Networks (GCN) has inspired a lot of methods~\cite{yan2018spatial,liu2020disentangling,channelwise2021,Enhancement2022,li2019spatio,Dynamic2020,MotifGCNs2022,Constructing2022} to exploit spatial and temporal dependencies among the joints by aid of the topology of human skeletons.
The pioneer work ST-GCN~\cite{yan2018spatial} learns local patterns within body parts by performing spatial-temporal graph convolutions on skeleton graphs built upon physical connections of joints.
Then the following works boost the graph modeling ability by exploring potential dependencies among physically apart joints with extended adjacency relations~\cite{liu2020disentangling,channelwise2021,Enhancement2022} and self-attention mechanisms~\cite{li2019spatio,Dynamic2020,MotifGCNs2022,Constructing2022}.
Although the above methods have achieved promising results on high-quality skeletons, the limitation in robustness against skeleton noise stands in the way of applying them in real tasks.

Recently, a few studies have considered low-quality skeletons and made efforts to learn noise-robust models.
A typical solution is to regard this task as a skeleton denoising problem.
Upon the human physiological constraints, linear denoising transformations~\cite{liu2017enhanced,nie2019view} are explored to filter linear coordinate variations in pose estimates but fail to address potential noise caused by unknown nonlinear transformations.
To tackle these issues, Demisse et al.~\cite{demisse2018pose} separately modeled the linear and nonlinear components of skeleton noise and employ an autoencoder to learn the nonlinear transformation.
However, it is hard to recover the corrupted skeletons through independent denoising operations as the skeleton degradation mechanism is unpredictable.
Another group of works~\cite{song2020richly,yoon2022predictively,song2022learning,bian2021structural} boost the model robustness against various degradations from the perspective of feature enhancement rather than skeleton correction.
Song et al.~\cite{song2020richly} developed a multi-stream architecture, where an individual stream focuses on a group of skeleton joints. 
Integration of multiple streams can produce redundant but complementary features.
Obviously, the heavy networks lead to low model efficiency, which is always an important consideration in practical applications.
Yu et al.~\cite{yoon2022predictively} trained a robust model by encoding the information shared between paired high-quality and low-quality skeletons of the same action instance in the global feature space.
Song et al.~\cite{song2022learning} made full use of unpaired skeletons and adapted global feature embeddings of low-quality skeletons towards the feature space of high-quality skeletons in an adversarial learning manner.
Bian et al.~\cite{bian2021structural} developed a graph matching scheme to transfer structural knowledge from high-quality skeletons to low-quality skeletons by aid of the common pose topology.
Compared to models using global features~\cite{yoon2022predictively,song2022learning}, the work~\cite{bian2021structural} achieves finer feature enhancement at joint-level but requires paired samples of high-quality and low-quality skeletons with homogeneous topology.
In this paper, we propose a more general and flexible model which enables part-level knowledge transfer between heterogeneous high-quality and low-quality skeletons.
The proposed model also relieves the requirement of paired high-quality and low-quality skeletons, thereby can incorporate solitary low-quality skeletons into training.

\subsection{Knowledge distillation}

Knowledge distillation is first proposed for model compression~\cite{hinton2015distilling}, in which a lightweight student is trained to mimic a larger teacher to achieve comparable performance.
Then this teacher-student structure is employed in some other scenarios~\cite{2018Complete,ge2020efficient}.
With respect to skeleton-based action recognition, there have been some attempts to construct high-performance students by distilling various forms of knowledge (e.g., output logits~\cite{thoker2019cross,yang2022mke,zhuang2023time,liu2022novel} and intermediate features~\cite{wang2022skeleton,cheng2021extremely}) from teachers.
Thoker and Gall~\cite{thoker2019cross} utilized RGB-skeleton pairs to achieve knowledge transfer from RGB videos to skeleton sequences.
Some recent studies~\cite{yang2022mke,wang2022skeleton} exploit various skeleton modalities, such as joint and bone, to improve the action recognition performance.
Yang et al.~\cite{yang2022mke} proposed a multi-modal knowledge distillation strategy to train a multi-modal student under the supervision of multiple single-modal or multi-modal teachers.
Wang et al.~\cite{wang2022skeleton} considered another situation that multiple modalities are observed during training while only one modality is available for inference.
They proposed an adaptive cross-form learning paradigm which forces a single-modal model to learn useful features from models based on other modalities.
Besides the above cross-modal scenarios, knowledge distillation techniques are also utilized in model compression~\cite{cheng2021extremely,zhuang2023time} and early action recognition~\cite{liu2022novel}.
Cheng et al.~\cite{cheng2021extremely} developed a margin ReLU distillation technique that utilizes features produced by a heavy teacher to supervise the learning of a lightweight student.
Zhuang et al.~\cite{zhuang2023time} forced a lightweight student to mimic the logits derived from a high-cost teacher to transfer long-term temporal knowledge.
Liu et al.~\cite{liu2022novel} boosted the early action recognition by distilling prior knowledge at different progress levels from complete sequences.

Our work aims to construct a noise-robust action recognition model by distilling feature-based knowledge from available high-quality skeletons to low-quality ones in the training stage.
Unlike the previous work~\cite{thoker2019cross,yang2022mke,zhuang2023time,liu2022novel,wang2022skeleton,cheng2021extremely}, we achieve adaptive part-level feature transfer to learn important local patterns from high-quality skeletons.
Specifically, an action-specific high-efficiency part matrix is devised to direct the student's attention towards body parts that are distinctive for action identification.

\section{Our Method}

\subsection{Motivation and the overall framework}
This paper aims to address the problem of action recognition based on low-quality skeletons.
While low-quality skeletons can be obtained more efficiently, the presence of substantial noise, including missing and inaccurate skeleton joints, presents significant challenges for action identification.
To this end, we propose to construct noise-robust representations for low-quality skeletons with the help of part-level action knowledge transferred from high-quality skeletons.
In real-life scenarios, low-quality and high-quality skeletons acquired from different pose estimators can be represented by heterogeneous pose graphs.
Consequently, transferring knowledge joint by joint is impractical.
Despite potential differences in joint numbers and connections, these heterogeneous skeletons share common body parts.
Motivated by these observations, we regard joints within a body part as a whole and integrate joint features into part features.
Then knowledge transfer between heterogeneous skeletons can be achieved by adapting the representations of low-quality skeletons towards those of high-quality skeletons at part-level.

\begin{figure*}[t]
	\begin{center}
		\includegraphics[width=1\textwidth]{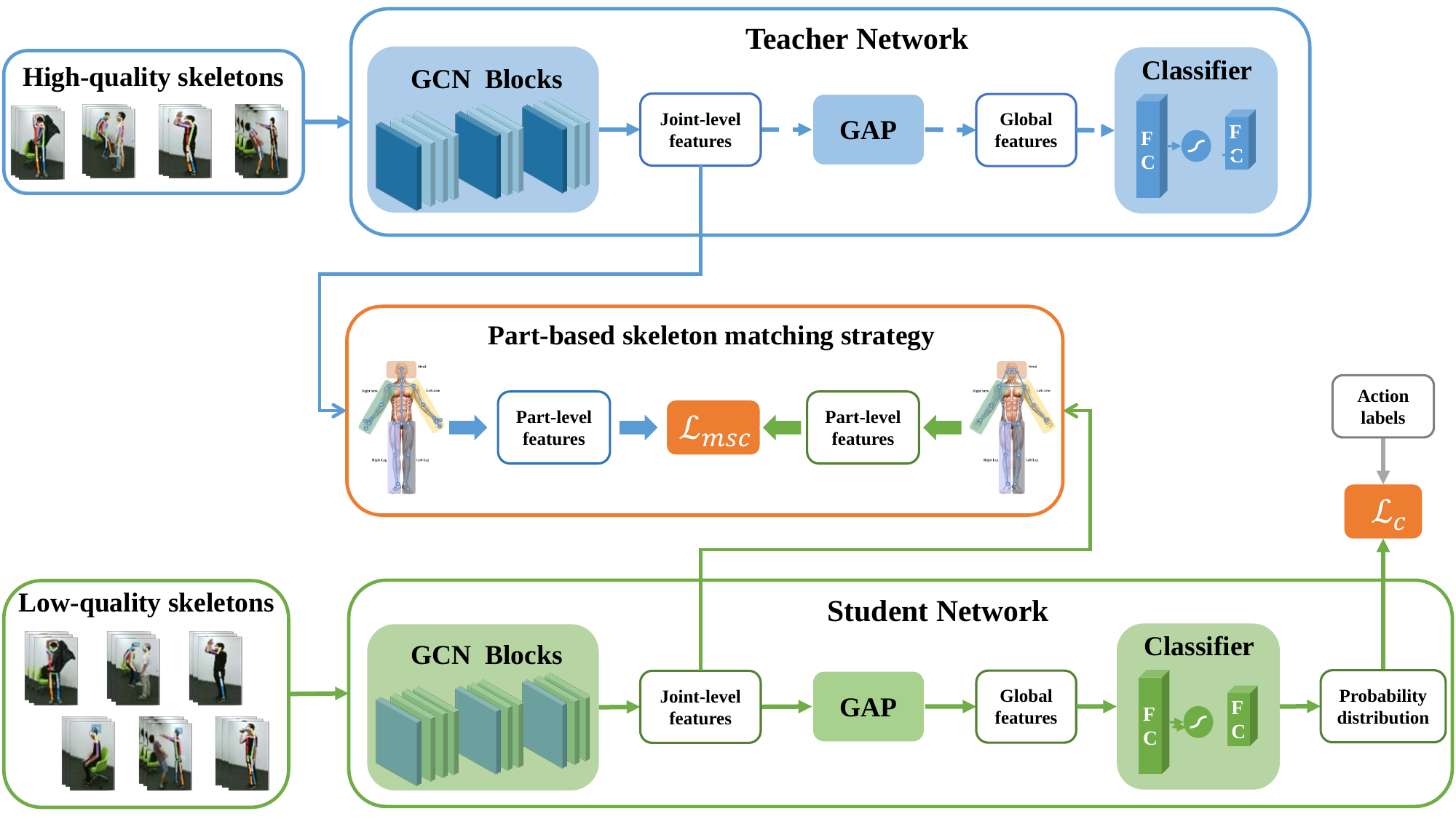}
	\end{center}
	\vskip -15pt
	\caption{Train pipeline of the proposed knowledge distillation framework for skeleton-based action recognition. It includes two processes: pre-training of teacher network (dashed lines) and training of student network (solid lines).}
	\vskip -10pt
	\label{fig:framework}
\end{figure*}

This study considers a general scenario that there exists a collection of training low-quality skeletons, with some having corresponding high-quality matches while others do not.
The training data consist of two components, namely paired skeletons $D_{pair}=\{(x_i^\mathrm{H},x_i^\mathrm{L},y_i)\}_{i=1}^M$ and solitary low-quality skeletons $D_{sol}=\{(x_i^\mathrm{L},y_i)\}_{i=M+1}^{M+N}$.
Here, $x_i^\mathrm{H}$ and $x_i^\mathrm{L}$ in $D_{pair}$ indicate high-quality and low-quality skeletons of an action instance, respectively.
$y_i \in \{1,2,...,C\}$ is the annotated action label and $C$ is the number of actions.
A low-quality skeleton sequence $x_i^\mathrm{L}$ in $D_{sol}$ has no corresponding high-quality data of the same action instance, but we can enhance their representations with prior knowledge provided by other high-quality skeletons.

The proposed method adopts the teacher-student framework, which provides a paradigm to train the student model with prior knowledge from a teacher model.
The teacher model is pre-trained on high-quality skeletons in $D_{pair}$, while the student model is learned on all accessible low-quality skeletons in $D_{pair}$ and $D_{sol}$.
Both the student model and the teacher model are built upon Graph Convolution blocks (GCN blocks) that construct representations of each joint by taking advantage of the skeleton topology graph.
We group the skeleton joints into five parts and take part features as the prior knowledge to guide the training of the student model.
A novel part-based skeleton matching strategy is developed to adaptively transfer discriminative knowledge from key parts of different actions.
By doing this, the student model can receive fine-grained guidance from the teacher, especially when certain body parts are lost or inaccurate in the low-quality skeletons.
In order to enhance the representations of low-quality skeletons in $D_{sol}$, the student model learns both intra-class and inter-class knowledge from other high-quality skeletons with a part-level multi-sample contrastive loss.
Fig.~\ref{fig:framework} depicts the overall architecture of the proposed method.

\begin{figure*}[t]
	\begin{center}
		\includegraphics[width=1\textwidth]{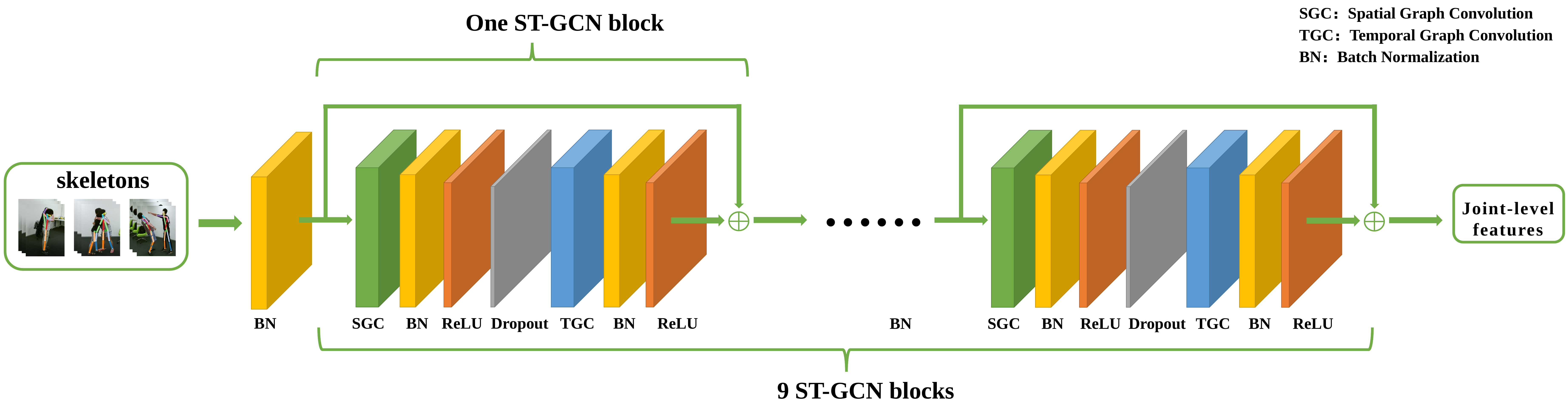}
	\end{center}
	\vskip -15pt
	\caption{Architecture of the GCN blocks built upon ST-GCN~\cite{yan2018spatial}. }
	\vskip -10pt
	\label{fig:GCNstructure}
\end{figure*}

\subsection{Teacher network}
A teacher network is trained on accessible high-quality skeletons $\{(x_i^\mathrm{H},y_i)\}_{i=1}^M$.
The teacher network contains a series of spatio-temporal GCN blocks for feature extraction, a Global Average Pooling (GAP) operation for information aggregation, and a C-way classifier for action identification.

Fig.~\ref{fig:GCNstructure} illustrates the detailed architecture of the GCN blocks built upon ST-GCN~\cite{yan2018spatial}.
Nine GCN blocks are employed in our implementation, alternately performing spatial graph convolution operation and temporal graph convolution operation to update the feature embeddings of joints.
Specifically, the spatial graph convolution operation integrates spatial context of neighboring joints in the skeleton graph and the temporal graph convolution operation aggregates the context of adjacent frames.
Note that the proposed knowledge distillation framework is also applicable to other advanced GCN blocks in addition to ST-GCN.
Suppose that a high-quality skeleton sequence includes $V$ joints, the operation of GCN blocks can be expressed as
\begin{equation}
	z_i^\mathrm{H} = \mathcal{F}(x_i^\mathrm{H}),
\end{equation}
where $z_i^\mathrm{H} \in R^{C \times T\times V}$ stands for the generated joint-level features.
$C$ indicates the number of channels and $T$ denotes the number of output temporal steps.
Usually, $T$ is equal to or smaller than the length of input frames.

With a global average pooling layer, the learned features $z_i^\mathrm{H}$ are integrated to a $C$-dimensional global vector, which is fed to the classifier for action recognition.
The classifier outputs the predicted probability distribution $P_i^\mathrm{H}$ for a skeleton sequence $x_i^\mathrm{H}$.
Then parameters of the teacher network are optimized by minimizing the cross entropy between the one-hot action label vector $Y_i$ and the predicted probability distribution $P_i^\mathrm{H}$.
The classification loss $\mathcal{L}_{teacher}$ is formulated by
\begin{equation}
	\mathcal{L}_{teacher}=-\frac{1}{M} \sum_{i=1}^M \sum_{c=1}^C Y_{i,c} \log P_{i, c}^\mathrm{H},
	\label{Eq:lossTeacher}
\end{equation}
where $Y_{i, c}=1$ if the true label is the $c$-th action, and $Y_{i, c}=0$ otherwise.
$P_{i, c}^\mathrm{H}$ indicates the predicted probability of $x_i^\mathrm{H}$ belonging to the $c$-th action.

\subsection{Student network}

With the pre-trained teacher network, the student network is trained under the guidance of the ground-truth labels and additional knowledge distilled from high-quality skeletons.
To achieve knowledge transfer between heterogeneous skeletons, we propose a novel part-based skeleton matching strategy which employs a part-level multi-sample contrastive loss to adapt part features of low-quality skeletons to that of high-quality skeletons.
As shown in Fig.~\ref{fig:part}, low-quality skeletons and high-quality skeletons share common parts in spite of their heterogeneous pose graphs.
Therefore, part features of high-quality skeletons can be regarded as local action knowledge to realize part-to-part knowledge transfer.
Compared to joint-level knowledge transfer that works on homogeneous skeletons, part-level knowledge transfer excels in capturing the inherent local patterns that often manifest through coordinated movements of neighboring skeleton joints.
On the other hand, part-level knowledge can provide fine-grained guidance for the student model to deal with partial occlusion.

\subsubsection{Part features}
The student network takes all available low-quality skeletons in $D_{pair}$ and $D_{sol}$ as input.
Similar to the teacher network, the student network also employs a series of GCN blocks (see Fig.~\ref{fig:GCNstructure}) to extract the feature embeddings.
This procedure can be formulated as
\begin{equation}
	z_i^\mathrm{L} = \mathcal{G}(x_i^\mathrm{L}),
\end{equation}
where $\mathcal{G}(\cdot)$ is the mapping function parameterized with multiple GCN blocks and $z_i^\mathrm{L} \in R^{C \times T\times V'}$ denotes the generated joint-level features of a low-quality skeleton sequence.
Note that the number of joints $V'$ can be different from that of high-quality skeletons.
In order to obtain part features, skeleton joints are grouped into five parts (i.e., head, left arm, right arm, left leg, and right leg) according to the physical connections of human body.
Then the joint-level features $z_i^\mathrm{L}$ are integrated into five part features $\{f_{i,p}^\mathrm{L} \in R^{C}\}_{p=1:5}$ by performing an average pooling on features of joints within each body part.
Similarly, five part feature vectors $\{f_{i,p}^\mathrm{H} \in R^{C}\}_{p=1:5}$ are derived from the original features $z_i^\mathrm{H}$ of a high-quality skeleton sequence $x_i^\mathrm{H}$.

\subsubsection{Training loss}

With the part-level knowledge distilled from high-quality skeletons, the student network is learned under the guidance of a classification loss $\mathcal{L}_{c}$ and a part-level multi-sample contrastive loss $\mathcal{L}_{pmsc}$.
The former guides the student network to classify low-quality skeletons to the correct actions and the latter enforces part feature alignment between low-quality and high-quality skeletons.
The total loss for optimizing the student network is formulated by
\begin{equation}
	\mathcal{L}_{student} = \mathcal{L}_{c} + \alpha * \mathcal{L}_{pmsc},
\end{equation}
where $\alpha$ is a weight balancing the two components.

The classifier in the student network takes the global feature vector of a sequence as input and predicts the probability distributions of actions.
The classification loss $\mathcal{L}_{c}$ is calculated by minimizing the cross entropy between the predicted probabilities and the action annotations:
\begin{equation}
	\mathcal{L}_{c}=-\frac{1}{(M+N)} \sum_{i=1}^{M+N} \sum_{c=1}^C Y_{i,c} \log P_{i, c}^\mathrm{H}.
	\label{Eq:lossCE}
\end{equation}
For an input skeleton sequence $x_i^\mathrm{L}$, $Y_{i}$ denotes the one-hot vector generated from its ground-truth label $y_i$ and $Y_{i,c}=1$ if $y_i=c$.
$P_{i, c}^\mathrm{H}$ indicates the predicted probability of the skeleton sequence belonging to the $c$-th action.

Our part-level multi-sample contrastive loss $\mathcal{L}_{pmsc}$ distills prior knowledge from multiple high-quality skeletons with the objective of decreasing intra-class variations across low-quality and high-quality skeletons while enlarging inter-class variations.
To compute $\mathcal{L}_{pmsc}$, we first employ high-quality skeletons in a mini-batch to construct a positive sample set $B^\mathrm{P}_{i}$ and a negative sample set $B^\mathrm{N}_{i}$ for a low-quality skeleton sequence $x_i^\mathrm{L}$.
To be specific, a mini-batch consists of a low-quality batch $B^\mathrm{L}$ and a high-quality batch $B^\mathrm{H}$.
Skeletons in $B^\mathrm{L}$ and $B^\mathrm{H}$ are sent to the student network and the teacher network to obtain part features, respectively.
For a low-quality skeleton sequence $x_i^\mathrm{L} \in B^\mathrm{L}$, we regard the positive sample set as $B^\mathrm{P}_{i} = \{x_j^\mathrm{H} \in B^\mathrm{H}| y_j=y_i, i\neq j\}$ and the negative sample set as $B^\mathrm{N}_{i} = \{x_j^\mathrm{H} \in B^\mathrm{H}| y_j \neq y_i\}$.
It should be noted that, the training low-quality data $\{x_i^\mathrm{L}\}_{i=1}^{M+N}$ come from the paired skeleton set $D_{pair}$ and the solitary skeleton set $D_{sol}$.
Especially, for a low-quality skeleton sequence $x_i^\mathrm{L}\in D_{pair}$ (i.e., $i \leq M$), there exists a high-quality skeleton sequence $x_i^\mathrm{H}$ extracted from the same action instance.

Next, we utilize $x_i^\mathrm{H}$, $B^\mathrm{P}_{i}$, and $B^\mathrm{N}_{i}$ in the part-level multi-sample contrastive loss.
Concretely, the similarity $\mathcal{C}(x_i^\mathrm{L},x_j^\mathrm{H})$ between a pair of low-quality skeleton sequence and high-quality skeleton sequence is given by
\begin{equation}
	\mathcal{C}(x_i^\mathrm{L},x_j^\mathrm{H}) = e^{\phi(\{f_{i,p}^\mathrm{L}\}, \{f_{j,p}^\mathrm{H}\})},
\end{equation}
\begin{equation}
	\phi(\{f_{i,p}^\mathrm{L}\}, \{f_{j,p}^\mathrm{H}\})=\sum_{p=1}^P (\frac{f_{i,p}^\mathrm{L} \cdot f_{j,p}^\mathrm{H}}{|f_{i,p}^\mathrm{L}| \cdot |f_{j,p}^\mathrm{H}|} \cdot E_{y_i, p}),
	\label{Eq:similarity}
\end{equation}
where $\phi(. ,.)$ is a similarity metric that measures the correlation of body parts from $x_i^\mathrm{L}$ and $x_j^\mathrm{H}$ by using their part features $\{f_{i,p}^\mathrm{L}\}$ and $\{f_{j,p}^\mathrm{H}\}$.
$E \in R^{C \times 5}$ is the action-specific high-efficiency part matrix and $E_{c,p} \in [0,1]$ indicates the efficient value of body part $p$ to the $c$-th action.
It is clear that the matrix $E$ makes body parts contribute differently to the correlation measurement in terms of the ground-truth action $y_i$ of $x_i^\mathrm{L}$.
Construction of the action-specific high-efficiency part matrix is explained in detail in Sec.~\ref{Sec:matrix}.
Then the knowledge distillation loss corresponding to $x_i^\mathrm{L}$ can be calculated by
\begin{equation}
	\ell^i=
	\begin{cases}
		-\log \frac{\mathcal{C}(x_i^\mathrm{L},x_i^\mathrm{H}) + w \cdot \mathcal{C}^{P}(x_i^\mathrm{L}) }
		{\mathcal{C}(x_i^\mathrm{L},x_i^\mathrm{H}) + w \cdot \mathcal{C}^{P}(x_i^\mathrm{L}) + \mathcal{C}^{N}(x_i^\mathrm{L})}, & x_i^\mathrm{L}\in D_{pair} \\
		-\log \frac{w \cdot \mathcal{C}^{P}(x_i^\mathrm{L}) }
		{w \cdot \mathcal{C}^{P}(x_i^\mathrm{L}) + \mathcal{C}^{N}(x_i^\mathrm{L})}, & x_i^\mathrm{L}\in D_{sol} \\
	\end{cases}
	\label{Eq:ell}
\end{equation}
\begin{equation}
	\mathcal{C}^{P}(x_i^\mathrm{L})=\frac{\sum_{x_j^\mathrm{H} \in B_{i}^P}\mathcal{C}(x_i^\mathrm{L},x_j^\mathrm{H}) }{|B_{i}^P|},
\end{equation}
\begin{equation}
	\mathcal{C}^{N}(x_i^\mathrm{L})=\frac{\sum_{x_j^\mathrm{H} \in B_{i}^N}\mathcal{C}(x_i^\mathrm{L},x_j^\mathrm{H}) }{|B_{i}^N|}.
\end{equation}
$\mathcal{C}^{P}(x_i^\mathrm{L})$ denotes the mean similarity between $x_i^\mathrm{L}$ and high-quality skeletons in $B_{i}^P$ and $\mathcal{C}^{N}(x_i^\mathrm{L})$ represents the mean similarity between $x_i^\mathrm{L}$ and skeletons in $B_{i}^N$.
The pre-defined parameter $w$ is a weight of $\mathcal{C}^{P}(x_i^\mathrm{L})$ and empirically set to $[0,1]$.
By minimizing $\ell^i$, we adapt $x_i^\mathrm{L}$ to high-quality skeletons of the same class, while pushing it away from high-quality skeletons of other classes.
Especially, for a low-quality skeleton sequence $x_i^\mathrm{L}$ in $D_{pair}$, we make a distinction between its corresponding high-quality skeleton sequence $x_i^\mathrm{H}$ and other positive skeletons in $B^\mathrm{P}_{i}$ to highlight the importance of $x_i^\mathrm{H}$ in knowledge distillation.
When $x_i^\mathrm{L}$ comes from $D_{sol}$, the student network can also transfer useful information from high-quality skeletons in $B_{i}^P$ and $B_{i}^N$ to construct robust features.
Finally, the knowledge distillation loss $\mathcal{L}_{pmsc}$ is calculated as the average value of all training low-quality skeleton sequences.
\begin{equation}
	\mathcal{L}_{pmsc}=\frac{1}{(M+N)} \sum_{i=1}^{M+N} \ell^i.
	\label{Eq:lpmsc}
\end{equation}

\subsubsection{Action-specific high-efficiency part matrix} \label{Sec:matrix}
In general, an action is jointly performed by all the skeleton joints, but the contribution of different joints to the recognition of different actions varies greatly.
Inspired by this, we explore an action-specific high-efficiency part matrix $\mathrm{E}\in R^{C\times 5}$ to capture the inherent relationships between body parts and actions.
With this matrix, the student model can adaptively emphasize prior knowledge of body parts that are critical to recognizing the input action instance as shown in Eq.~\ref{Eq:similarity}.

In order to construct the action-specific high-efficiency part matrix, we utilize the pre-trained teacher model and high-quality training skeleton sequences $\{x_i^\mathrm{H}\}_{i=1:M}$ to evaluate the contribution of each body part.
Given a skeleton sequence $x_i^\mathrm{H}$, five occlusion skeleton sequences $\{x_{i,p}^\mathrm{H}\}_{p=1:5}$ are derived from it by occluding the joints within a certain body part.
Next, the occlusion skeleton sequences are fed to the pre-trained teacher model for action identification.
Suppose that the ground-truth label of an occlusion skeleton sequence $x_{i,p}^\mathrm{H}$ is $y_{i,p}$ and the predicted label is $\hat y_{i,p}$, we evaluate the efficient value $\mathrm{E}_{c,p}$ of a body part $p$ for the $c$-th action by calculating the ratio of mis-classified instances.
More specifically,
\begin{equation}
	\mathrm{E}_{c,p}=\frac{\sum_{i=1}^M \mathds{1}(y_{i,p}=c, \hat y_{i,p} \neq c)} { \sum_{i=1}^M \mathds{1}(y_{i,p}=c)},
\end{equation}
where $\mathds{1}(\cdot)$ is an indicator function, that is $\mathds{1}(\cdot)=1$ if the pre-defined expression is true, otherwise $\mathds{1}(\cdot)=0$.
Apparently, a large efficient value $\mathrm{E}_{c,p}$ means that body part $p$ plays a critical role in recognizing action $c$ and occluding this part would result in a high mis-classification rate.
On the contrary, a small efficient value $\mathrm{E}_{c,p}$ indicates that part $p$ is trivial for identifying action $c$.
Finally, efficient values of parts are normalized by the softmax operation:
\begin{equation}
	\mathrm{E}_{c, p}=\frac{\exp ^{E_{c, p} }}{\sum_{p=1}^5 \exp ^{E_{c, p} }}.
\end{equation}

\begin{figure*}[t]
	\begin{center}
		\includegraphics[width=1\textwidth]{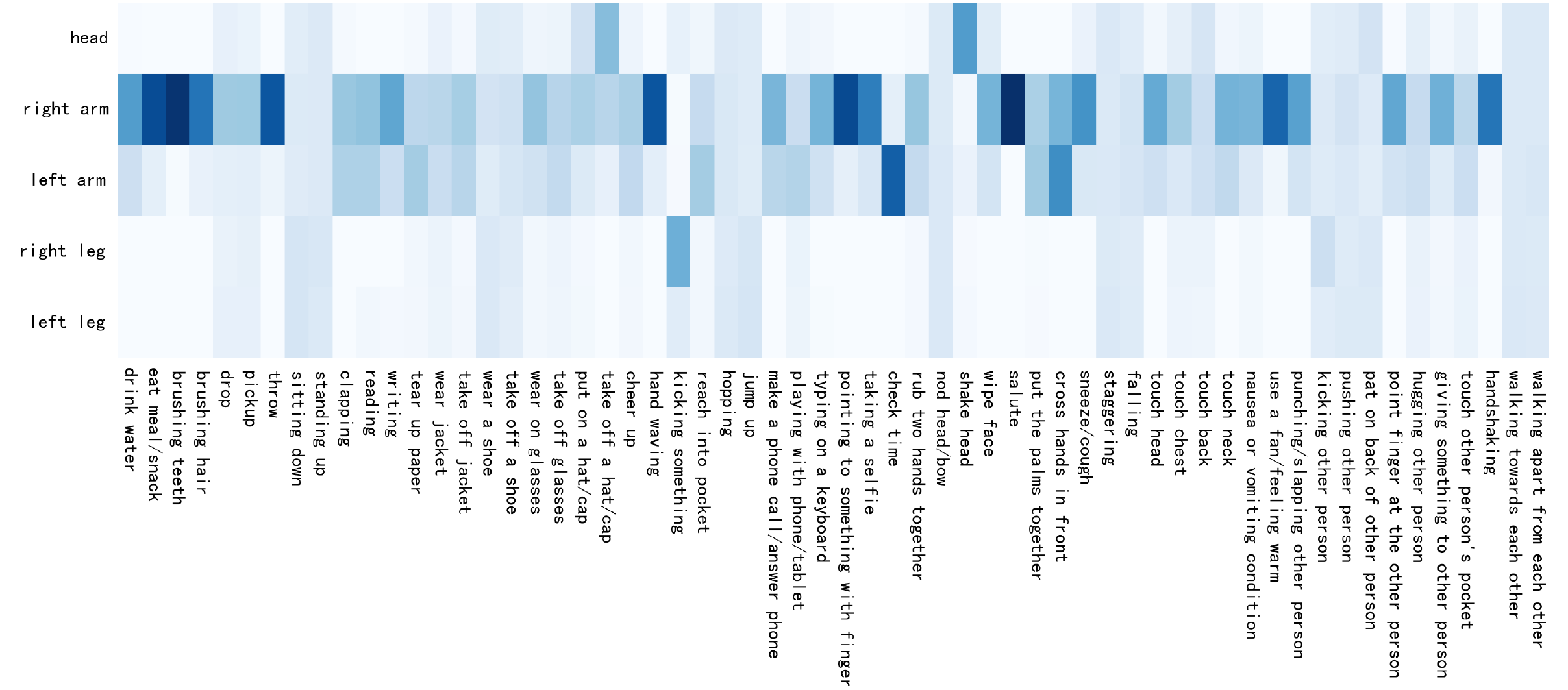}
	\end{center}
	\vskip -10pt
	\caption{Visualization of the action-specific high-efficiency part matrix on the NTU-RGB+D dataset. Large values are represented in dark blue, while small values are expressed in pale. This figure is best seen in color.}
	\vskip -10pt
	\label{fig:matrix-ntu60}
\end{figure*}

Fig.~\ref{fig:matrix-ntu60} visualizes the action-specific high-efficiency part matrix learned on the NTU-RGB+D dataset~\cite{shahroudy2016ntu}.
The learned matrix basically conforms to the characteristics of various actions.
For example, joints of arms are of great importance to identify actions such as ``eat meal/snack" and ``drink water", while joints of legs play a critical role in recognizing actions such as ``kicking something" and ``kicking other person".
For actions that involve the body cooperative movement, such as ``falling" and ``walking towards each other", multiple parts are assigned similar efficient values.

\section{Experiments}

\subsection{Datasets} We evaluate the proposed method on three large-scale action recognition datasets: the NTU-RGB+D dataset~\cite{shahroudy2016ntu}, the Penn Action dataset~\cite{zhang2013actemes} and the SYSU 3D HOI dataset~\cite{hu2015jointly}.

The NTU-RGB+D dataset~\cite{shahroudy2016ntu} employs three Microsoft Kinect V.2 sensors to capture a diverse set of 60 distinct actions performed by 40 subjects from various viewpoints. 
These actions encompass 50 single-person actions and 10 interactions involving two individuals, resulting in a comprehensive dataset of 56,880 action instances.
Each action instance is available in four data formats, including RGB videos, depth image sequences, 3D skeleton sequences, and infrared image sequences.
Authors of the dataset proposed two evaluation benchmarks: the Cross-Subject (CS) benchmark and the Cross-View (CV) benchmark. 
For the CS benchmark, the training set comprises action instances performed by 20 subjects, while the testing set contains data from the remaining 20 subjects.
In the case of the CV benchmark, action instances captured by camera 2 and camera 3 are utilized for training, while those recorded by camera 1 are exclusively reserved for testing.

The Penn Action dataset~\cite{zhang2013actemes} comprises 2,326 action instances, distributed across 15 distinct action categories. 
Each action instance provides two types of data modalities: RGB videos and manually annotated skeleton sequences. 
This dataset is partitioned into the training set and the testing set, each containing 1,163 samples.
We employ the Top1 and Top5 accuracies on the testing set as evaluation metrics in this section.

The SYSU 3D HOI dataset~\cite{hu2015jointly} focuses on human-object interactions and comprises 480 video clips involving 12 actions performed by 40 subjects. 
For each action instance, the subject interacts with one of six objects: a telephone, a chair, a bag, a wallet, a mop, or a broom.
Authors of the dataset have provided two different settings for model evaluation. 
In setting 1, half of the instances for each action are used for training, while the remaining half are used for testing. 
In setting 2, actions performed by half of the subjects are used for training purpose, and the actions from the remaining subjects are used for testing. 
There is no overlap of subjects between the training and testing sets.
For each setting, the dataset provides 30 random train/test splits. 
We report the average accuracy across these 30 distinct splits to give a comprehensive assessment of the proposed method.

\subsection{Skeleton data}
This work aims to develop a high-performance action recognition model based on low-quality skeletons with significant noise. 
In the training stage, the model leverages available high-quality skeletons to distill valuable knowledge and learn noise-robust representations for low-quality skeletons.
Specifically, the NTU-RGB+D dataset provides 3D coordinates of 25 joints captured by Kinect V.2 sensors. 
The SYSU 3D HOI dataset records actions using Kinect V.1 sensors, generating 3D coordinates of 20 joints.  
The Penn Action dataset offers manually annotated skeleton data composed of 13 joints. Additionally, we employ the PifPaf pose estimation algorithm~\cite{kreiss2019pifpaf} to extract 2D coordinates and confidences for 17 body joints from RGB videos. 
Note that, the estimated 2D skeletons have heterogeneous skeleton graphs to the skeleton data offered by the datasets.
To obtain skeletons with varying qualities, we adopt three different backbones: ResNet152~\cite{he2016deep}, ShuffleNetv2x1~\cite{ma2018shufflenet}, and MobileNetv3Small~\cite{howard2019searching}.
Among them, ResNet152 generates more accurate skeletons but comes with a computational cost that is more than twice that of ShuffleNetv2x1 and MobileNetv3Small. 
Generally, 2D skeletons generated using the ResNet152 backbone and the skeletons provided by the datasets are of relatively high quality, while 2D skeletons generated using ShuffleNetv2x1 or MobileNetv3Small backbones tend to be of lower quality.
In the following experiments, we evaluate the proposed method under different settings.

\subsection{Implementation details}
The proposed method adopts GCN blocks built upon ST-GCN~\cite{yan2018spatial}.
The number of subjects for each action instance is set to 2, and the video frame length is set to 300. 
For action instances performed by a single subject or those with a frame length less than 300, padding with zeros is employed to achieve the specified size.
Excess video frames are discarded if the video frame length exceeds 300.
The weight $w$ in Eq.~\ref{Eq:ell} is empirically set to 0.5.
During the training phase, the model is optimized using the SGD algorithm with a momentum of 0.9 and weight decay of $10^{-4}$. 
The learning rate is initialized to 0.1 and is decayed by a factor of 0.1 at epochs 30, 40, and 50. 
The training process stops after completing 60 epochs.

\subsection{Experimental results}
\subsubsection{Knowledge distillation from heterogeneous high-quality skeletons}
In this setting, a student model takes low-quality skeletons as input and transfers knowledge from a teacher model pre-trained on high-quality skeletons represented by a heterogeneous pose graph.
On the NTU-RGB+D dataset, skeletons produced by Microsoft Kinect V.2 are regarded as high-quality data while skeletons assessed by the PifPaf algorithm with ShuffleNetv2x1 as the backbone are taken as low-quality data.
On the Penn Action dataset, high-quality skeletons are those annotated by human experts and low-quality skeletons are generated by the PifPaf algorithm, using ResNet152 as the backbone.
On the SYSU 3D HOI dataset, the PifPaf algorithm, when employing ResNet152 as the backbone network, can produce more accurate skeleton data than Microsoft Kinect V.1.
Therefore, skeletons assessed using the PifPaf algorithm and those generated by Microsoft Kinect V.1 are considered as high-quality and low-quality skeletons, respectively.
Body occlusion is a natural problem in real-world scenarios, which results in absence of some joints.
To validate the robustness of the proposed model, we additionally simulate the degradation of low-quality skeletons by introducing random occlusions to each joint with a specified probability.
If a joint is selected to be occluded, it's coordinates are set to zero.

\begin{table*}[t]
	\centering
	\caption{Experimental results of the proposed method by distilling knowledge from \textbf{heterogeneous} high-quality skeletons. Student (w/o KD) indicates a model learned on low-quality skeletons without distilling knowledge from the teacher. Student (w/ KD) represents a student model that distills part-level knowledge from the teacher. KD is short for Knowledge Distillation. OP is short for Occluded Probability.}
	\label{tab:heterogeneous}
	\scalebox{0.68}{
		\begin{tabular}{c|cc|cc}\toprule[2pt]
			\textbf{Methods} & \multicolumn{2}{c|} {\textbf{Data Settings}} & \multicolumn{2}{c}{\textbf{Accuracy (\%)}}\\ 
			\midrule[1.2pt]
			\textbf{NTU-RGB+D} & \textbf{Skeleton} & \textbf{OP} & \textbf{CS} & \textbf{CV} \\ \hline
			Teacher & Microsoft Kinect V.2 & 0 & 84.56 & 89.46 \\
			Student (w/o KD) & PifPaf-ShuffleNetv2x1 & 0 & 79.98 & 84.88 \\
			Student (w/ KD) & PifPaf-ShuffleNetv2x1 & 0 & \textbf{83.31(+3.33)} & \textbf{88.13(+3.25)}   \\ 
			Student (w/o KD) & PifPaf-ShuffleNetv2x1 & 0.3 & 76.30 & 80.45  \\
			Student (w/ KD) & PifPaf-ShuffleNetv2x1 & 0.3 & \textbf{81.76(+5.46)} & \textbf{85.67(+5.22)} \\ 
			Student (w/o KD) & PifPaf-ShuffleNetv2x1 & 0.6 & 73.25 & 77.70  \\
			Student (w/ KD) & PifPaf-ShuffleNetv2x1 & 0.6 & \textbf{79.41(+6.16)} & \textbf{84.03(+6.33)}  \\ 
			\midrule[1.2pt]
			\textbf{Penn Action} & \textbf{Skeleton} & \textbf{OP} & \textbf{Top1} & \textbf{Top5} \\ \hline
			Teacher & Human & 0 & 90.54  & 98.97   \\
			Student (w/o KD) & PifPaf-ResNet152 & 0 & 83.43  & 97.19  \\
			Student (w/ KD) & PifPaf-ResNet152 & 0 & \textbf{87.08(+3.65)} & \textbf{98.50(+1.31)} \\ 
			Student (w/o KD) & PifPaf-ResNet152 & 0.3 & 78.18&  96.63\\
			Student (w/ KD) & PifPaf-ResNet152 & 0.3 &  \textbf{84.74(+6.56)} & \textbf{98.31 (+1.68)}\\ 
			Student (w/o KD) & PifPaf-ResNet152 & 0.6 & 70.51&  95.13\\
			Student (w/ KD) & PifPaf-ResNet152 & 0.6 & \textbf{80.52(+10.01)} & \textbf{96.82 (+1.69)}\\ 
			\midrule[1.2pt]
			\textbf{SYSU 3D HOI} & \textbf{Skeleton} & \textbf{OP} & \textbf{Setting1} & \textbf{Setting2} \\ \hline			
			Teacher & PifPaf-ResNet152 & 0 & 88.54  & 88.96   \\
			Student (w/o KD) & Microsoft Kinect V.1 & 0 & 84.33  & 84.00 \\
			Student (w/ KD) & Microsoft Kinect V.1 & 0 & \textbf{87.14(+2.81)} & \textbf{86.68(+2.68)} \\ 
			Student (w/o KD) & Microsoft Kinect V.1 & 0.3 & 77.95&  78.35\\
			Student (w/ KD) & Microsoft Kinect V.1 & 0.3 &  \textbf{81.43(+3.48)} & \textbf{81.76(+3.41)}\\ 
			Student (w/o KD) & Microsoft Kinect V.1 & 0.6 & 65.39&  66.01\\
			Student (w/ KD) & Microsoft Kinect V.1 & 0.6 & \textbf{68.88(+3.49)} & \textbf{68.41(+2.40)}\\ 
			\bottomrule[2pt]
		\end{tabular}
	}
\vskip -10pt
\end{table*}

Experimental results on three datasets are depicted in Table~\ref{tab:heterogeneous}.
Teacher models undergo training and evaluation using high-quality skeletons, while student models are trained and tested on low-quality skeletons.
The proposed method ( Student (w/ KD) in Table~\ref{tab:heterogeneous} ) is compared with a baseline method ( Student (w/o KD) ) which is also trained using low-quality skeletons, but without the knowledge distillation from the teacher model.
We can obtain the following observations from Table~\ref{tab:heterogeneous}.
\begin{itemize}
\item We can see that the quality of skeletons has a large impact on the performance of skeleton-based action recognition.
Take the NTU-RGB+D dataset for example, the action recognition accuracy of teacher on CS benchmark is 84.56\%, while it drops to 79.98\% by using low-quality skeletons.
When the model utilizes synthesized low-quality skeletons with occluded probability 0.3 and 0.6, the accuracy further decreases to 76.30\% and 73.25\%, respectively.
\item  Although high-quality and low-quality skeletons are represented by heterogeneous pose graphs, the proposed method can distill part-level knowledge from high-quality skeletons to boost the action recognition performance based on low-quality ones. 
As a result, the student model using low-quality skeletons can attain performance comparable to that of the teacher model learned on high-quality skeletons.
\end{itemize}

\subsubsection{Knowledge distillation from homogeneous high-quality skeletons}

In this setting, the proposed method is evaluated by using homogeneous high-quality and low-quality skeletons produced by the PifPaf algorithm.
Across the three datasets, ResNet152 serves as the backbone network for producing high-quality skeletons.
To investigate the robustness of the proposed method, we employ two backbone networks to generate low-quality skeletons.
Specifically, ShuffleNetv2x1 is selected as the backbone network on the NTU-RGB+D dataset, while a more lightweight network, MobileNetv3Small, is adopted as the backbone for the Penn Action and SYSU 3D HOI datasets.
Table~\ref{tab:homogeneous} summarizes the experimental results on the three datasets.

On the NTU-RGB+D dataset, the teacher model pre-trained on skeletons produced by PifPaf-ResNet152 achieves 84.65\% and 90.41\% recognition accuracy on CS and CV benchmarks, respectively.
When using ShuffleNetv2x1 as the backbone, the generated skeletons include more inaccurate joints and the action recognition accuracy of the student model drops to 79.98\% (CS) and 84.88\% (CV).
Similar results can be observed on the Penn Action and SYSU 3D HOI datasets.
With the help of part-level knowledge distillation from high-quality skeletons, performance of the student model can be significantly improved, which demonstrates that the proposed knowledge distillation framework is also effective for homogeneous high-quality and low-quality skeletons.
In general, the higher the occluded probability of low-quality skeletons, the more performance improvement the student can obtain through knowledge distillation. 
This observation further validates the robustness of the proposed method against absent joints caused by partial occlusion or data missing in signal transmission.

\begin{table*}[t]
	\centering
	\caption{Experimental results of the proposed method by distilling knowledge from \textbf{homogeneous} high-quality skeletons. Student (w/o KD) indicates a model learned on low-quality skeletons without distilling knowledge from the teacher. Student (w/ KD) represents a student model that distills part-level knowledge from the teacher. KD is short for Knowledge Distillation. OP is short for Occluded Probability.}
	\label{tab:homogeneous}
	\scalebox{0.68}{
		\begin{tabular}{c|cc|cc}\toprule[2pt]
			\textbf{Methods} & \multicolumn{2}{c|} {\textbf{Data Settings}} & \multicolumn{2}{c}{\textbf{Accuracy (\%)}}\\ 
			\midrule[1.2pt]
			\textbf{NTU-RGB+D} & \textbf{Skeleton} & \textbf{OP} & \textbf{CS} & \textbf{CV} \\ \hline
			Teacher & PifPaf-ResNet152 & 0 & 84.65 & 90.41  \\
			Student (w/o KD) & PifPaf-ShuffleNetv2x1 & 0 & 79.98 & 84.88  \\
			Student (w/ KD) & PifPaf-ShuffleNetv2x1 & 0 & \textbf{82.50(+2.52)}  & \textbf{88.36(+3.48)} \\ 
			Student (w/o KD) & PifPaf-ShuffleNetv2x1 & 0.3 & 76.30 & 80.45  \\
			Student (w/ KD) & PifPaf-ShuffleNetv2x1 & 0.3 & \textbf{81.58(+5.28)} & \textbf{86.61(+6.16)}  \\ 
			Student (w/o KD) & PifPaf-ShuffleNetv2x1 & 0.6 & 73.25 & 77.70  \\
			Student (w/ KD) & PifPaf-ShuffleNetv2x1 & 0.6 & \textbf{80.09(+6.84)}  & \textbf{83.61(+5.91)} \\ 
			\midrule[1.2pt]
			\textbf{Penn Action} & \textbf{Skeleton} & \textbf{OP} & \textbf{Top1} & \textbf{Top5} \\ \hline
			Teacher & PifPaf-ResNet152 & 0 & 83.43& 97.19   \\
			Student (w/o KD) & PifPaf-MobileNetv3Small & 0 &73.88  &95.90   \\
			Student (w/ KD) & PifPaf-MobileNetv3Small & 0 &\textbf{78.84(+4.96)}  &\textbf{97.10(+1.20)} \\ 
			Student (w/o KD) & PifPaf-MobileNetv3Small & 0.3 & 69.10  &  94.52 \\
			Student (w/ KD) & PifPaf-MobileNetv3Small & 0.3 &\textbf{74.91(+5.81)}  &\textbf{96.63(+2.11)} \\ 
			Student (w/o KD) & PifPaf-MobileNetv3Small & 0.6 & 63.30  & 91.29  \\
			Student (w/ KD) & PifPaf-MobileNetv3Small & 0.6 &\textbf{73.22+(9.92)}  & \textbf{95.79(+4.50)}\\ 
			\midrule[1.2pt]
			\textbf{SYSU 3D HOI} & \textbf{Skeleton} & \textbf{OP} & \textbf{Setting1} & \textbf{Setting2} \\ \hline			
			Teacher & PifPaf-ResNet152 & 0 & 88.54  & 88.96   \\
			Student (w/o KD) & PifPaf-MobileNetv3Small & 0 &86.39  &86.67   \\
			Student (w/ KD) & PifPaf-MobileNetv3Small & 0 &\textbf{87.58(+1.19)}  &\textbf{88.17(+1.50)} \\ 
			Student (w/o KD) & PifPaf-MobileNetv3Small & 0.3 & 81.15  &  82.43 \\
			Student (w/ KD) & PifPaf-MobileNetv3Small & 0.3 &\textbf{84.07(+2.92)}  &\textbf{84.67(+2.24)} \\ 
			Student (w/o KD) & PifPaf-MobileNetv3Small & 0.6 & 70.40  & 70.32  \\
			Student (w/ KD) & PifPaf-MobileNetv3Small & 0.6 &\textbf{80.29(+9.89)}  &\textbf{81.14(+10.82)}\\ 
			\bottomrule[2pt]
		\end{tabular}
	}
	\vskip -10pt
\end{table*}

\subsection{Visualization of action recognition results}
\begin{figure}[t]
	\centering
	\includegraphics[width=0.88\linewidth]{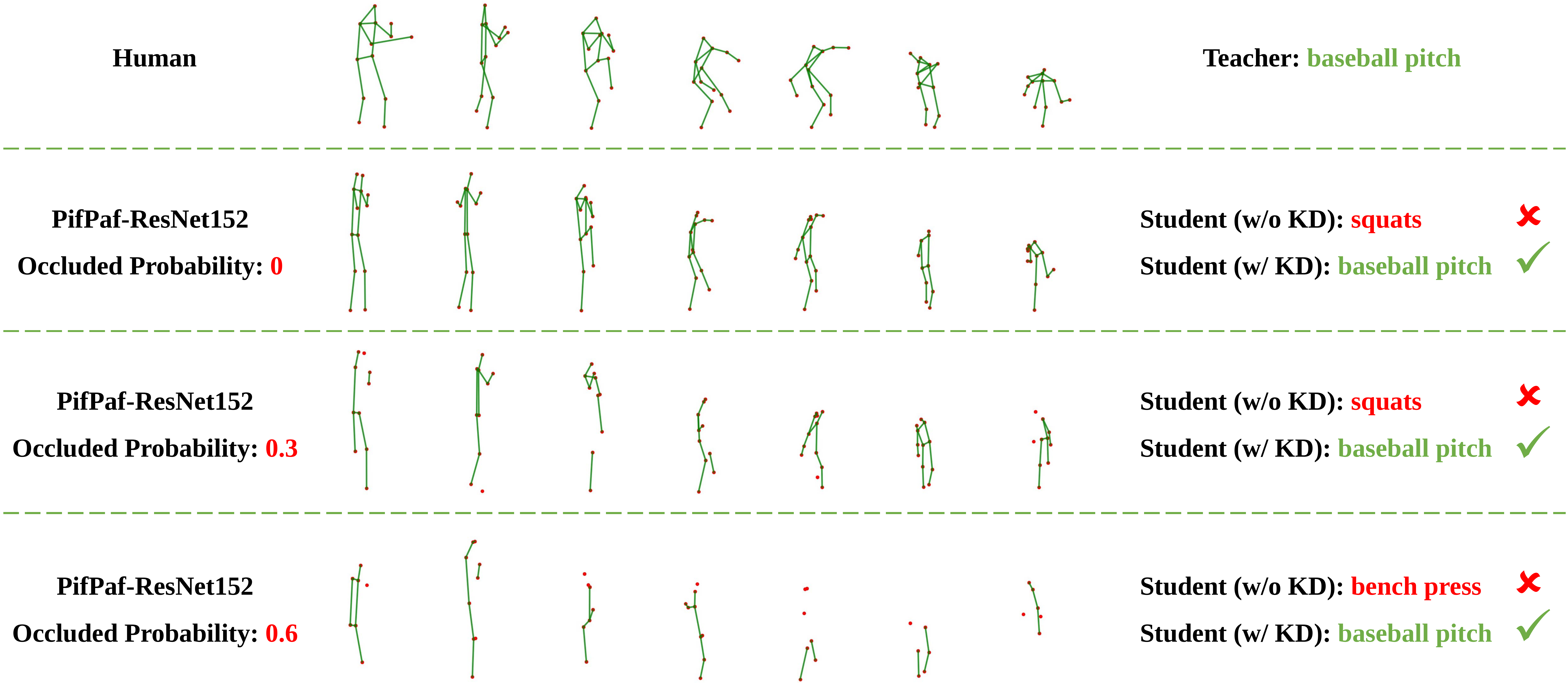}
	\vskip -5pt
	\caption{Visualization of the action recognition results on the Penn Action dataset.}
	\label{fig:vis-Penn-hetero}
	\vskip -0pt
\end{figure}

To further investigate the effectiveness of knowledge distillation from heterogeneous skeletons, Fig.~\ref{fig:vis-Penn-hetero} visualizes the skeleton data and action recognition results of one instance on the Penn Action dataset.
The teacher model is trained on high-quality skeletons annotated by human experts, while the student model takes low-quality skeletons generated by the PifPaf algorithm with ResNet152 as the backbone.
We can clearly observe that manually annotated skeletons provide a more accurate description of the pose changes, but skeletons extracted through the PifPaf algorithm exhibit noticeable missing joints and positional offsets, due to the impact of viewpoints and self-occlusions.
As the occlusion probability increases from 0 to 0.6, the missing joints in low-quality skeletons become increasingly noticeable, making action recognition more challenging.
However, even in the presence of significant noise, the proposed method (i.e., Student (w/ KD)) can still learn discriminative representations through knowledge distillation, allowing for accurate recognition of action categories.

\begin{figure}[t]
	\centering
	\includegraphics[width=0.88\linewidth]{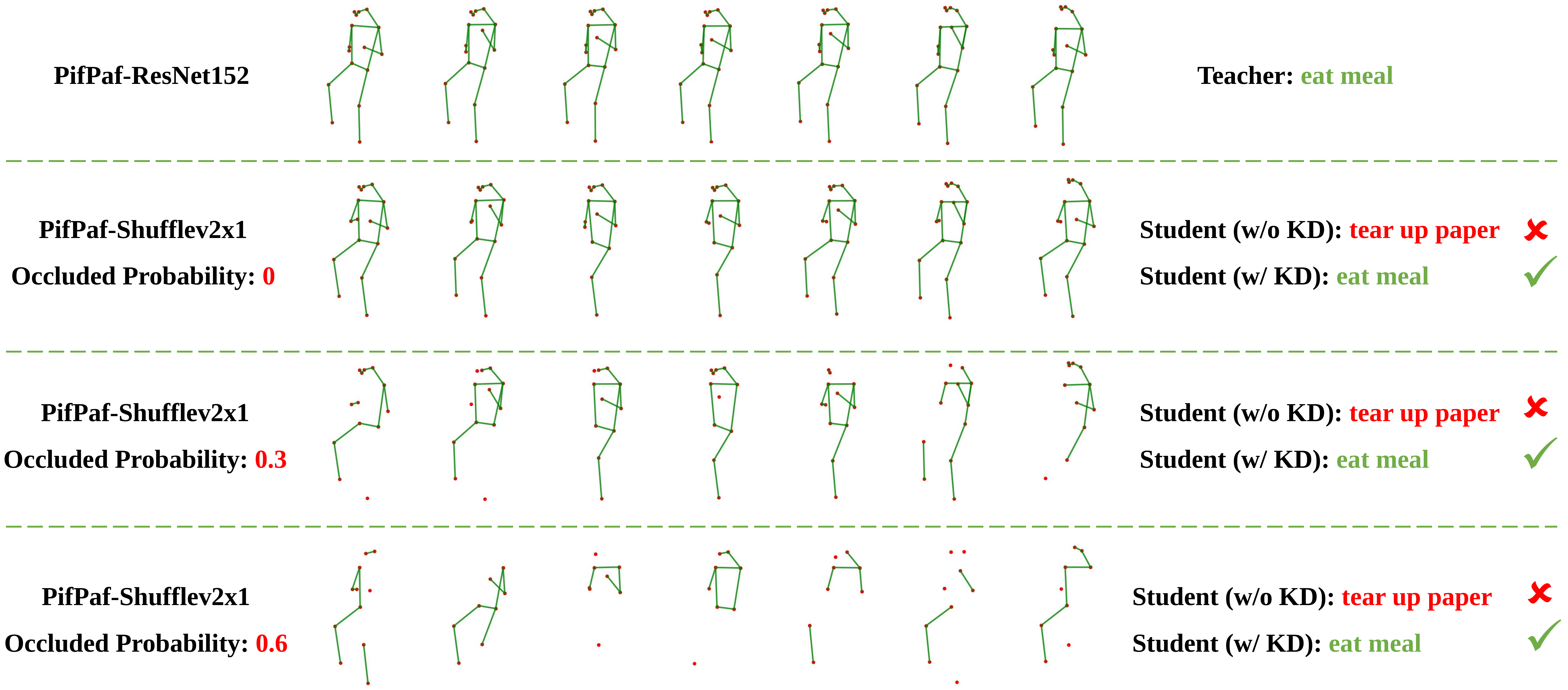}
	\vskip -5pt
	\caption{Visualization of the action recognition results on the NTU-RGB+D dataset.}
	\label{fig:vis-NTU-homo}
	\vskip -10pt
\end{figure}

Fig.~\ref{fig:vis-NTU-homo} depicts the action recognition results of an ``eat meal" instance on the NTU-RGB+D dataset.
This figure also visualizes homogeneous high-quality and low-quality skeletons derived from the PifPaf algorithm by using different backbone networks.
In high-quality skeletons, two hand joints are positioned with a certain horizontal distance, representing the action of one-handed eating.
However, two hand joints are closely spaced in low-quality skeletons, causing the model to incorrectly  recognize it as the action of ``tear up paper".
Transferring part-level knowledge from high-quality skeletons to low-quality skeletons allows the proposed model to acquire more precise local motion patterns, enabling accurate action identification even for instances with missing joints.

\begin{table*}[t]
	\centering
	\caption{Comparison with the existing action recognition methods. KD stands for Knowledge Distillation.}
	\label{tab:compare}
	\scalebox{0.68}{
		\begin{tabular}{c|cc|cc}  \toprule[2pt]
			\textbf{Methods} & \multicolumn{2}{c|} {\textbf{Settings}} & \multicolumn{2}{c}{\textbf{Accuracy (\%)}}\\ 
			\midrule[1.2pt]
			\textbf{NTU-RGB+D} & \textbf{Skeleton} & \textbf{GCN Blocks} & \textbf{CS} & \textbf{CV} \\ 
			\hline
			3S-RA-GCN~\cite{song2020richly} & PifPaf-ShuffleNetv2x1 & ST-GCN  & 76.35 & 82.75 \\
			SKD~\cite{bian2021structural} & PifPaf-ShuffleNetv2x1& ST-GCN & 80.80 &  84.42  \\ 
			PE-GCN~\cite{yoon2022predictively} & PifPaf-ShuffleNetv2x1& ST-GCN & 80.22 &  84.91 \\
			PE-GCN~\cite{yoon2022predictively} & PifPaf-ShuffleNetv2x1& Js-AGCN & 81.20 &  88.74  \\ 
			PR-GCN~\cite{li2021pose} & PifPaf-ShuffleNetv2x1& -- & 80.70 & 87.83  \\
			\textbf{Ours (KD)} & PifPaf-ShuffleNetv2x1& ST-GCN & \textbf{83.31} & \textbf{88.13}  \\ 
			\textbf{Ours (KD)}  & PifPaf-ShuffleNetv2x1 & Js-AGCN &\textbf{83.25} &  \textbf{90.14}\\
			\hline
			PA-ResGCN-B19~\cite{song2020stronger} &PifPaf-ShuffleNetv2x1 &-- &86.49   &91.27 \\
			EfficientGCN-B4~\cite{Constructing2022} &PifPaf-ShuffleNetv2x1 &-- &87.57  &91.81 \\
			\textbf{Ours (KD)} & PifPaf-ShuffleNetv2x1 &EfficientGCN-B4  &\textbf{88.36}  &\textbf{92.12}  \\ 
			\midrule[1.2pt]
			\textbf{Penn Action} & \textbf{Skeleton} & \textbf{GCN Blocks} & \textbf{Top1} & \textbf{Top5} \\
			\hline
			3S-RA-GCN~\cite{song2020richly} & PifPaf-ResNet152 & ST-GCN  & 77.43&95.60 \\
			SKD~\cite{bian2021structural} & PifPaf-ResNet152 & ST-GCN &  81.87	&95.44 \\
			PE-GCN~\cite{yoon2022predictively} & PifPaf-ResNet152 & Js-AGCN &75.09  &94.85    \\ 
			PR-GCN~\cite{li2021pose} & PifPaf-ResNet152 & -- &87.89 &98.24  \\
			\textbf{Ours (KD)} & PifPaf-ResNet152 & ST-GCN &\textbf{87.08} &\textbf{98.50}  \\ 
			\textbf{Ours (KD)}  & PifPaf-ResNet152 & Js-AGCN &\textbf{90.45}  &\textbf{98.69}  \\
			\hline
			PA-ResGCN-B19~\cite{song2020stronger} &PifPaf-ResNet152 &-- &92.79   &97.21 \\
			EfficientGCN-B4~\cite{Constructing2022} &PifPaf-ResNet152 &-- &93.45  &97.84 \\
			\textbf{Ours} & PifPaf-ResNet152 &EfficientGCN-B4  &\textbf{95.03}  &\textbf{98.22}  \\ 
			\midrule[1.2pt]
			\textbf{SYSU 3D HOI} & \textbf{Skeleton} & \textbf{GCN Blocks} & \textbf{Setting1} & \textbf{Setting2} \\ 
			\hline
			3S-RA-GCN~\cite{song2020richly} &PifPaf-MobileNetv3Small & ST-GCN  & 80.04 & 79.60 \\
			PE-GCN~\cite{yoon2022predictively} &PifPaf-MobileNetv3Small & ST-GCN & 85.23 &  85.08 \\
			PE-GCN~\cite{yoon2022predictively} &PifPaf-MobileNetv3Small & Js-AGCN & 85.33 &  85.92  \\ 
			PR-GCN~\cite{li2021pose} &PifPaf-MobileNetv3Small & -- & 87.29 & 88.46  \\
			\textbf{Ours (KD)} &PifPaf-MobileNetv3Small & ST-GCN & \textbf{87.58} & \textbf{88.17}  \\ 
			\textbf{Ours (KD)}   &PifPaf-MobileNetv3Small & Js-AGCN &\textbf{88.31} &  \textbf{89.14}\\
			\hline
			PA-ResGCN-B19~\cite{song2020stronger} &PifPaf-MobileNetv3Small &-- &86.79   &88.21 \\
			EfficientGCN-B4~\cite{Constructing2022} &PifPaf-MobileNetv3Small &-- &88.68  &89.31 \\
			\textbf{Ours (KD)} & PifPaf-MobileNetv3Small &EfficientGCN-B4  &\textbf{90.07}  &\textbf{91.00}  \\
			\bottomrule[2pt]
		\end{tabular}
	}
	\vskip -10pt
\end{table*}

\subsection{Comparison with the existing methods}
We compare the proposed method with previous skeleton-based action recognition methods and present the experimental results on the NTU-RGB+D, Penn Action, and SYSU 3D HOI datasets in Table~\ref{tab:compare}.
All the compared methods employ the same low-quality skeletons for both training and evaluation.
These skeletons are explicitly denoted in the ``skeleton" column of Table~\ref{tab:compare}.
On the NTU-RGB+D dataset, our method distills knowledge from heterogeneous high-quality skeletons produced by Microsoft Kinect V.2.
On the Penn Action dataset, manually annotated skeletons serve as heterogeneous high-quality data for our method.
On the SYSU 3D HOI dataset, our method learns from homogeneous high-quality skeletons generated by the PifPaf algorithm with Resnet152 as the backbone network.

Among the comparison methods listed in Table~\ref{tab:compare}, 3S-RA-GCN~\cite{song2020richly}, SKD~\cite{bian2021structural}, PE-GCN~\cite{yoon2022predictively}, and PR-GCN~\cite{li2021pose} are end-to-end networks specifically designed to handle low-quality skeletons, eliminating the need for a separate denoising stage.
Concretely, 3S-RA-GCN~\cite{song2020richly} employs a multi-stream architecture to learn redundant yet noise-robust features, at the cost of reduced model efficiency.
PE-GCN~\cite{yoon2022predictively} encodes the key information shared between normal skeletons and noisy skeletons by using global features.
SKD~\cite{bian2021structural} achieves joint-level knowledge distillation between low-quality and high-quality skeletons.
PR-GCN~\cite{li2021pose} introduces a pose refinement module to correct the joint coordinates before feeding them into a two-branch GCN, which fuses both position and motion information of joints.
As shown in Table~\ref{tab:compare}, our method outperforms these methods by using the same low-quality skeletons for training and evaluation.
Furthermore, we also notice that replacing the GCN blocks from ST-GCN~\cite{yan2018spatial} to Js-AGCN~\cite{shi2019two} leads to an improvement in action recognition accuracy, which validates the generality of our method across various GCN-based models.

In Table~\ref{tab:compare}, our proposed method is also compared to two State-Of-The-Art (SOTA) models for skeleton-based action recognition, namely PA-ResGCN-B19~\cite{song2020stronger} and EfficientGCN-B4~\cite{Constructing2022}.
In this group of experiments, our proposed method employs EfficientGCN-B4 as the GCN blocks to extract joint-level features.
It is noteworthy that Table~\ref{tab:compare} presents the performance of PA-ResGCN-B19 and EfficientGCN-B4 on low-quality skeleton data, revealing a distinct decrease compared to their original performance on high-quality skeleton data reported in~\cite{song2020stronger} and~\cite{Constructing2022}. 
This disparity underscores the substantial challenges inherent in action recognition based on low-quality skeleton data.
Our proposed method, utilizing part-level knowledge distillation, demonstrates superior performance over SOTA models~\cite{song2020stronger,Constructing2022} on the same low-quality skeletons across various datasets.

\subsection{Ablation studies}
\subsubsection{Evaluation of the part-based skeleton matching strategy}

In this section, we evaluate the proposed part-based skeleton matching strategy, which utilizes a novel part-level multi-sample contrastive loss ($\mathcal{L}_{pmsc}$ in Eq.~\ref{Eq:lpmsc}) to distill part-level knowledge from high-quality skeletons. 
To this end, we compare the proposed method with three baselines on the NTU-RGB+D dataset and summarize the action recognition results in Table~\ref{tab:baseline}.
All the baselines are derived from the student model by using different loss functions.
(1) The first baseline is a student model learned on low-quality skeletons without distilling knowledge from high-quality skeletons. 
Only the classification loss $\mathcal{L}_{c}$ in Eq.~\ref{Eq:lossCE} is utilized to train the network.
(2) Besides the classification loss, the second baseline introduces KL loss to mimic global features produced by the teacher.
(3) The third baseline achieves global feature-based knowledge transfer using a Global Multi-Sample Contrastive (Global MSC) loss~\cite{sharma2021instance}.
Different from the proposed part-based skeleton matching strategy, the similarities between low-quality and high-quality skeletons are measured on global features instead of part features.

\begin{table*}[t]
	\centering
	\caption{Ablation study results on the NTU-RGB+D dataset for evaluating the effectiveness of the part-based skeleton matching strategy.}	\label{tab:baseline}
	\scalebox{0.72}{
		\begin{tabular}{c|ccc|cc} \toprule[2pt]
			\multirow{2}{*}{\textbf{Methods}} & \multirow{2}{*}{\textbf{Skeleton}} &\multirow{2}{*}{\textbf{Knowledge distillation}} & \multirow{2}{*}{\textbf{Loss function}} & \multicolumn{2}{c}{\textbf{Accuracy (\%)}}\\ \cline{5-6}
			&&&& \textbf{CS} & \textbf{CV} \\
			\midrule[1pt]
			Student  & PifPaf-ShuffleNetv2x1 & -- & $\mathcal{L}_{c}$ &   79.98& 84.88  \\
			Student  & PifPaf-ShuffleNetv2x1 & Global feature-based KD & $\mathcal{L}_{c}$ + KL  & 80.17 & 86.74  \\
			Student  & PifPaf-ShuffleNetv2x1 & Global feature-based KD & $\mathcal{L}_{c}$ + Global MSC  & 81.95 & 86.11 \\
			Student  & PifPaf-ShuffleNetv2x1 & Part feature-based KD & $\mathcal{L}_{c}$ + $\mathcal{L}_{pmsc}$  &\textbf{83.31} & \textbf{88.13}  \\ 
			\bottomrule[2pt]
		\end{tabular}
	}
	\vskip -10pt
\end{table*}

The latter two baselines as well as the proposed method distill knowledge from a teacher model pre-trained on high-quality skeletons produced by the Microsoft Kinect V.2.
As shown in Table~\ref{tab:baseline}, the overall best results are achieved by knowledge distillation with the proposed part-based skeleton matching strategy.
The latter two baselines perform better than the first baseline, demonstrating that global feature-based knowledge distillation from high-quality skeletons can boost the performance of action recognition on low-quality data.
The global MSC loss tends to be more effective compared to the KL loss, as evidenced by the third baseline outperforming the second baseline on the CS benchmark and achieving comparable performance on the CV benchmark.
Compared to the third baseline using global MSC loss, the proposed part-based skeleton matching strategy can further improve the recognition accuracy of CS and CV benchmarks with a respective advantage of 1.36\% and 2.02\%.

\begin{table*}[t]\small
	\caption{Ablation study results (\%) on the NTU-RGB+D dataset by using partial high-quality skeletons.}
	\label{tab:general}
	\centering
	\scalebox{0.87}{
		\begin{tabular}{c|cc|cc}
			\toprule[2pt]
			\multirow{2}{*}{\textbf{Methods}} & \multicolumn{2}{c|}{\textbf{Data settings}} & \multicolumn{2}{c}{\textbf{Accuracy (\%)}}\\ \cline{2-5} 
			& \textbf{Skeleton} & \textbf{Amount} & \textbf{CS} & \textbf{CV} \\
			\midrule[1pt]
			Teacher & Microsoft Kinect V.2 & 80\% & 84.21 & 87.15 \\
			Student (w/o KD) & PifPaf-ShuffleNetv2x1 & 80\% & 77.12 & 82.97 \\
			Student (w/ KD)  & PifPaf-ShuffleNetv2x1 & 80\%  & 82.03 & 86.61 \\
			Student (w/ KD)  & PifPaf-ShuffleNetv2x1 & 100\% & \textbf{83.27(+1.24)} & \textbf{87.80(+1.19)} \\
			\midrule[1pt]		
			Teacher & Microsoft Kinect V.2 & 70\%  & 83.22 & 86.29  \\
			Student (w/o KD) & PifPaf-ShuffleNetv2x1 & 70\% & 76.07 & 81.85  \\
			Student (w/ KD)  & PifPaf-ShuffleNetv2x1 & 70\% & 80.94 & 85.69  \\
			Student (w/ KD)  & PifPaf-ShuffleNetv2x1 & 100\% & \textbf{82.61(+1.67)} & \textbf{87.36(+1.67)}  \\ 
			\midrule[1pt]
			Teacher & PifPaf-Resnet152 & 80\% & 83.48 & 90.28 \\
			Student (w/o KD) & PifPaf-ShuffleNetv2x1 & 80\% & 77.12 & 82.97\\
			Student (w/ KD) & PifPaf-ShuffleNetv2x1 & 80\% & 81.40  & 87.33  \\
			Student (w/ KD) & PifPaf-ShuffleNetv2x1 & 100\% & \textbf{82.24(+0.84)} & \textbf{87.99(+0.66)}   \\ 
			\midrule[1pt]
			Teacher & PifPaf-Resnet152 & 70\% & 82.08 & 88.27  \\
			Student (w/o KD) & PifPaf-ShuffleNetv2x1 & 70\% & 76.07 & 81.85  \\
			Student (w/ KD) & PifPaf-ShuffleNetv2x1 & 70\% & 80.65  & 86.06  \\
			Student (w/ KD) & PifPaf-ShuffleNetv2x1 & 100\%	& \textbf{81.31(+0.66)}  & \textbf{87.54(+1.48)}  \\ 
			\bottomrule[2pt]
		\end{tabular}
	}
	\vskip -10pt
\end{table*}

\subsubsection{Evaluation under more general settings}

This paper considers a more general case that part of the training low-quality skeletons has no corresponding high-quality matches extracted from the same action instance.
Four groups of experiments are conducted on the NTU-RGB+D dataset to evaluate the proposed method.
Low-quality skeletons are estimated by the PifPaf algorithm with ShuffleNetv2x1 as the backbone network.
We randomly select a certain proportion of action instances and acquire their corresponding high-quality skeletons to build the teacher model.
The first two groups adopt heterogeneous high-quality data produced by Microsoft Kinect V.2, while the last two groups utilize homogeneous high-quality data generated by the PifPaf algorithm with Resnet152 as the backbone network.
Experimental results are summarized in Table~\ref{tab:general}.

Take the first group for example, only 80\% of low-quality skeletons has their corresponding high-quality data and the learned teacher model achieves 84.21\% accuracy on CS benchmark.
By using 80\% of the low-quality data, the student model attains an accuracy of 77.12\% on the CS benchmark, which is improved to 82.03\% through knowledge distillation.
When incorporating the rest 20\% low-quality skeletons into training, the student model can enhance their representations through intra-class knowledge transfer facilitated by the part-level multi-sample contrastive loss.
As a result, the performance of the student model is further improved by 1.24\%.
Consistent findings are observed across the remaining three groups, demonstrating that the proposed knowledge distillation framework has good scalability towards individual low-quality skeletons with no corresponding high-quality data.

\section{Conclusion}
This paper has presented an innovative knowledge distillation framework that enhances action recognition using low-quality skeletons. 
By transferring fine-grained knowledge from high-quality to low-quality skeletons, the method effectively bridges the gap between different data qualities. 
A part-based skeleton matching strategy enables the extraction of local action patterns across heterogeneous pose graphs, and an action-specific high-efficiency part matrix further refines this process by highlighting crucial parts for accurate action identification.
The incorporation of a part-level multi-sample contrastive loss facilitates knowledge transfer for instances with only low-quality skeletons.
This work not only introduces a robust framework but also provides valuable insights for addressing challenges related to incomplete or inaccurate skeleton data in action recognition.
Extensive experiments on the NTU-RGB+D, Penn Action, and SYSU 3D HOI datasets have validated the effectiveness of the proposed knowledge distillation framework.

\section{Acknowledgment}
This work was supported in part by the Natural Science Foundation of Liaoning Province under Grant No.2021-MS-266, and in part by the Shenyang Young and Middle-aged Science and Technology Innovation Talent Support Program under Grant No.RC210427, and in part by the National Natural Science Foundation of China (NSFC) under Grant No.62171295, and in part by the Applied Basic Research Project of Liaoning Province under Grant 2023JH2/101300204.

\bibliographystyle{elsarticle-num}
\bibliography{reference-brief.bib}

\end{document}